\documentclass{comsoc2010}
\usepackage{amsmath}
\usepackage{amsthm}

\title{Where are the hard manipulation problems?}
\author{Toby Walsh}

\begin{document}
\newtheorem{theorem}{Theorem}
\newtheorem*{example}{Running Example:}
\newtheorem{definition}{Definition}
\newtheorem{myexample}{Example}
\newtheorem{mytheorem}{Theorem}
\newcommand{\myproof}{\noindent {\bf Proof:\ \ }}
\newcommand{\myqed}{\mbox{$\heartsuit$}}


\begin{abstract}
One possible escape from the Gibbard-Satterthwaite theorem is
computational complexity. For example, it is NP-hard to compute if
the STV rule can be manipulated. However, there is increasing concern
that such results may not reflect the difficulty of
manipulation in practice. In this tutorial, 
I survey recent results in this area.
\end{abstract}



The Gibbard Satterthwaite theorem proves that, under some
simple assumptions, a voting rule can always be manipulated. 
A number of possible escapes 
have been suggested.
For example, if we relax the 
assumption of an universal domain
and replace it with single peaked
preferences, then strategy free voting
rules exist. In an influential paper \cite{bartholditoveytrick},
Bartholdi, Tovey and Trick 
proposed that complexity might offer another escape:
perhaps it is computationally so difficult to find a successful
manipulation that agents have little
option but to report their true preferences? 
Many voting rules have subsequently
been shown to be NP-hard
to manipulate \cite{csljacm07}.
However, NP-hardness only dictates the worst-case
and may not reflect the difficulty of manipulation
in practice. Indeed, a number of recent theoretical results 
suggest that manipulation can often be easy (e.g. \cite{xcec08b}). 


I argue here that
we can study the hardness
of manipulation empirically \cite{wijcai09,wecai10}. 
There are several reasons why empirical
analysis is useful. 
For example, theoretical analysis is usually
restricted to simple distributions like uniform votes. 
Votes in real elections may be very different due, for instance, to 
correlations between votes. As a second example, theoretical 
analysis is often asymptotic so does not
reveal the size of hidden constants. Such
constants may be important to 
the actual computational cost. 
In addition, elections are typically bounded in
size so asymptotic results may be 
uninformative. 
%
%
Such experiments suggest
different behaviour occurs in the problem of 
computing manipulations
of voting rules
than in other NP-hard problems like
propositional satisfiability
\cite{cheeseman-hard,waaai98}, constraint satisfaction 
\cite{random,mpswcp98}, 
number partitioning 
\cite{rnp,gw-ci98}. 
and other NP-hard problems
\cite{GentIP:tsppt,wijcai99,wijcai2001}. 
For instance, many transitions seen in our experiments
appear smooth, as seen in polynomial problems 
\cite{waaai2002}.


Another problem in which manipulation
may be an issue is the stable marriage problem.
As with voting, an important issue is whether agents 
can manipulate the result by
mis-reporting their preferences. 
Unfortunately, Roth \cite{roth-manip} proved 
that {\em all} stable marriage procedures
can be manipulated. 
We might hope that 
computational complexity might also be a barrier
to manipulate stable marriage procedures. 
In joint work with
Pini, Rossi and Venable, I have proposed a new stable marriage
procedures based on voting that
is NP-hard to manipulate \cite{prvwaamas09}. 


A third domain in which manipulation
may be an issue is sporting tournaments
\cite{rwadt09}. 
Manipulating a sporting tournament is slightly
different to manipulating an election. In a sporting
tournament, the voters are also the candidates. 
Since it is hard (without bribery or similar mechanisms)
for a team to play better than it can, we consider
just manipulations where the manipulators can
throw games. 
We show, for example, that we can
decide how to manipulate round robin and cup competitions, two 
of the most popular sporting competitions 
in polynomial time. 


\bibliographystyle{plain}
\bibliography{/Users/twalsh/Documents/biblio/a-z,/Users/twalsh/Documents/biblio/pub,/Users/twalsh/Documents/biblio/a-z2,/Users/twalsh/Documents/biblio/pub2}



\begin{contact}
Toby Walsh\\
NICTA and UNSW\\
Sydney, Australia\\
\email{toby.walsh@nicta.com.au}
\end{contact}


\end{document}